\documentclass{article}

\usepackage[main, final]{neurips_2025}

\usepackage[utf8]{inputenc} 
\usepackage[T1]{fontenc}    
\usepackage{hyperref}       
\usepackage{url}            
\usepackage{booktabs}       
\usepackage{amsfonts}       
\usepackage{nicefrac}       
\usepackage{microtype}      
\usepackage{xcolor}         
\usepackage{tcolorbox} 
\usepackage{graphicx}
\usepackage{subcaption}
\usepackage{amsmath}

\title{MASCA: LLM-based Multi-Agent System for Credit Assessment}

\author{%
  Gautam Jajoo\thanks{Correspondence to: \href{mailto:jajoo@kairosity.ai}{\texttt{jajoo@kairosity.ai}}} \\
  Kairosity\\
  \And
  Atharva Pandey \\
  Kairosity \\
  \And
  Pranjal Chitale \\
  Microsoft Research \\
  \And
  Saksham Agarwal \\
  Independent Author \\
}

\begin{document}

\maketitle

\begin{abstract}
Recent advancements in financial problem-solving have leveraged LLMs and agent-based systems, with a primary focus on trading and financial modeling. However, credit assessment remains an underexplored challenge, traditionally dependent on rule-based methods and statistical models. In this paper, we introduce MASCA, an LLM-driven multi-agent system designed to enhance credit evaluation by mirroring real-world decision-making processes. The framework employs a layered architecture where specialized LLM-based agents collaboratively tackle sub-tasks. We further present a signaling game theory perspective on hierarchical multi-agent systems, offering theoretical insights into their structure and interactions. Our paper also includes a detailed bias analysis in credit assessment, addressing fairness concerns. Experimental results demonstrate that MASCA outperforms baseline approaches, highlighting the effectiveness of hierarchical LLM-based multi-agent systems in financial applications, particularly in credit scoring.
\end{abstract}

\section{Introduction}

The financial domain has witnessed a major shift with the introduction of Large Language Models (LLMs), which have demonstrated potential across various financial tasks. Recent studies have showcased the capabilities of advanced LLMs, such as GPT-4, in financial text analysis \cite{lopezlira2024chatgptforecaststockprice}, prediction tasks \cite{xie2023wallstreetneophytezeroshot}, and financial reasoning \cite{son2023classificationfinancialreasoningstateoftheart}. These models have proven particularly effective in processing and analyzing complex financial data, offering insights that were previously challenging to obtain through traditional methods.

Building upon the capabilities of LLMs, autonomous agents leveraging these models to tackle complex financial problems, have emerged as a powerful approach. Autonomous agents leverage LLMs to comprehend, generate, and reason with natural language, and this capability has been extended to the financial domain where they assist in tasks ranging from real-time market analysis to automated trading decisions \cite{xiao2025tradingagentsmultiagentsllmfinancial}. Such agents have shown promise not only in processing large volumes of financial data but also in engaging in strategic and collaborative decision-making. However, one area where their potential remains underexplored is credit assessment, a domain that requires processing diverse data sources and navigating dynamic borrower-lender interactions.

Traditional credit assessment and scoring methods, while widely used, face several critical challenges: they rely heavily on historical credit data, overlooking alternative data sources that could provide a more comprehensive view of creditworthiness. Historical data can also inadvertently perpetuate existing biases leading to unfair lending practices \cite{https://doi.org/10.1111/jofi.13090}. Traditional models operate as ``black boxes" in the decision-making processes of these systems, making it difficult to understand for consumers and regulators to interpret \cite{bracke2019machine}. Static models struggle to adapt quickly to changing economic conditions or evolving financial behaviors.

LLMs are uniquely positioned to address these challenges. Their ability to process unstructured and diverse data sources enables them to incorporate alternative data into credit assessments. Furthermore, their reasoning capabilities can enhance transparency by providing interpretable explanations for decisions. By integrating these models into a multi-agent framework, it becomes possible to create adaptive systems that respond dynamically to changing market conditions while promoting fairness and inclusivity.

This paper makes three contributions:
\begin{itemize}
    \item \textbf{LLM-based multi-agent framework for credit assessment}: improving accuracy, fairness, and adaptability in decision-making.
    \item \textbf{Hierarchical multi-agent structure with Signaling Game Theory}: capturing borrower-lender strategic interactions and information flow.
    \item \textbf{Bias analysis in LLM-based credit assessment}: identifying and mitigating systemic risks in financial decision-making.
\end{itemize}

\section{Related Work}
LLMs have demonstrated strong capabilities across various financial applications \cite{nie2024surveylargelanguagemodels}, such as analyzing sentiment in financial news and social media \cite{shen2024financialsentimentanalysisnews}, predicting market trends \cite{Fatouros_2024}, interpreting financial time series data \cite{yu2023temporaldatameetsllm,tang2024timeseriesforecastingllms}, and finding factors that influence stock movements \cite{wang2024llmfactorextractingprofitablefactors}. Their ability to extract relevant financial metrics and ratios from unstructured data has enhanced the speed and accuracy of financial assessments \cite{wang-brorsson-2025-large}. 

Multi-agent systems (MAS) have long been used in financial applications for their ability to model complex, dynamic environments \cite{Kampouridis_2022}, \cite{abuHakimaToloo1997}. Agents in these systems operate autonomously, interact with each other, and collaborate to achieve shared goals. MAS has been applied to tasks such as algorithmic trading, fraud detection, and dynamic portfolio management.

There has been previous research work on LLM-based agents such as FinMem \cite{yu2023finmemperformanceenhancedllmtrading}, a trading agent with layered memory to convert the insights gained from memories into investment decisions and FinAgent \cite{zhang2024multimodalfoundationagentfinancial}, which proposes a multimodal agent to reason for financial trading. Previous work on LLM-based multi-agent systems include financial decision-making \cite{yu2024finconsynthesizedllmmultiagent} and trading systems \cite{ding2024largelanguagemodelagent}, \cite{xiao2025tradingagentsmultiagentsllmfinancial}. 


\section{Methodology}

\begin{figure}[h]
    \centering
    \includegraphics[width=0.7\linewidth]{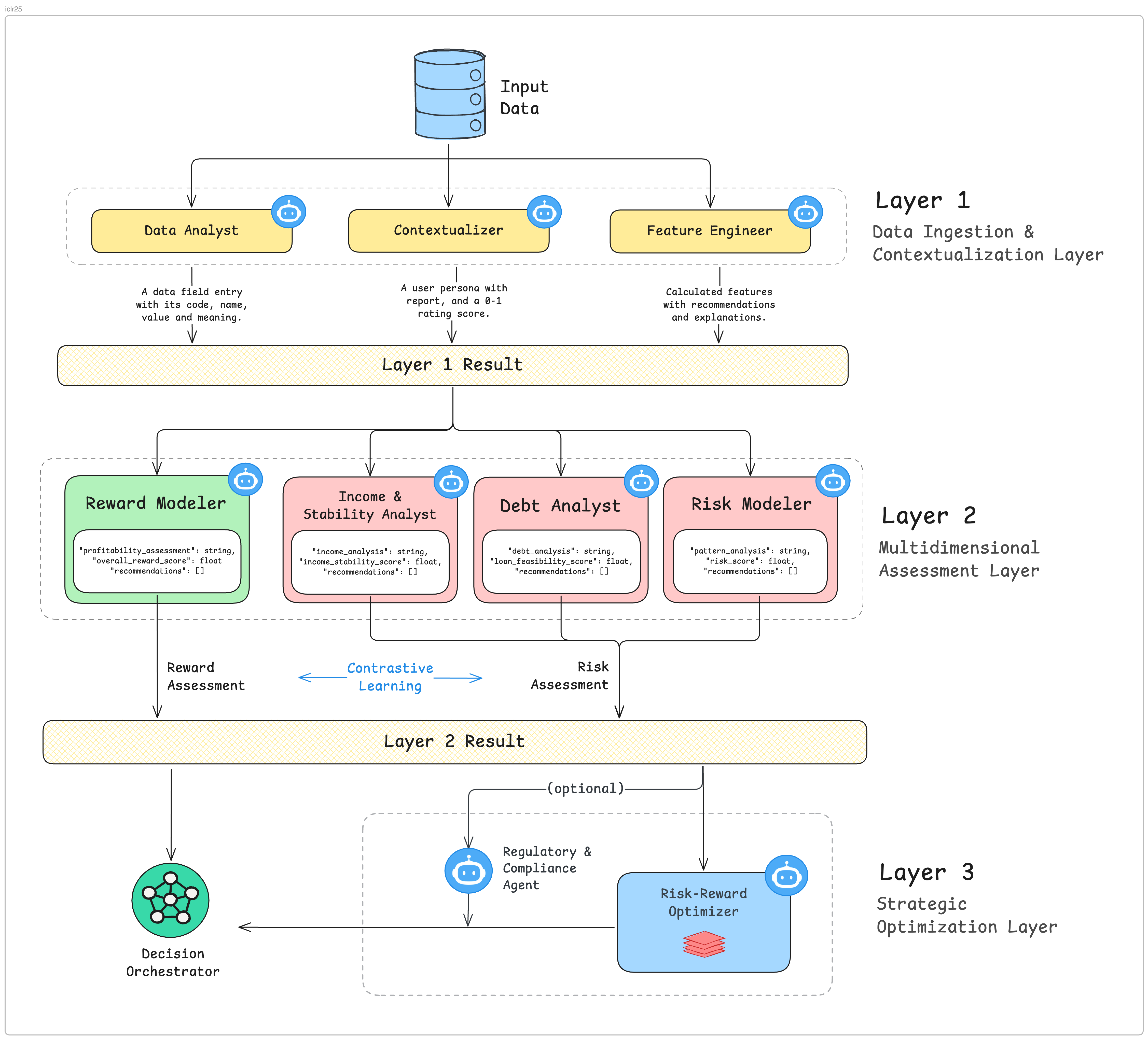}
    \caption{MASCA: The multi agent framework for credit assessment}
    \label{framework}
\end{figure}

Our hierarchical multi-agent system (MAS) for credit assessment (Figure \ref{framework}) mirrors real-world credit teams by decomposing the task into specialized, modular agents. This design ensures Modularity, Explainability, Specialization, and Scalability (MESS). Each layer focuses on distinct aspects of credit evaluation, enabling transparent, accurate, and efficient decision-making.

\textbf{Data Ingestion \& Contextualization Layer: }Transforms raw applicant data into structured profiles via three agents: \textbf{Data Analyst:} Aggregates, formats, and validates structured/unstructured data, \textbf{Contextualizer:} Builds applicant personas, integrating financial and behavioral insights, \textbf{Feature Engineer:} Derives key metrics (e.g., DTI, DAR, Credit Utilization, Employment Stability).

\textbf{Multidimensional Assessment Layer:} Conducts parallel evaluation of risks and rewards: The \textbf{Risk Team:} contains Risk Modeler (credit history, red flags), Income \& Stability Analyst (income consistency, employment history, stress tests), Debt Analyst (debt burden, loan specifics). The \textbf{Reward Modeler:} highlights profitability, creditworthiness, and mitigating factors.

\textbf{Strategic Optimization Layer \& Decision Orchestrator:} This layer contains \textbf{Risk-Reward Optimizer} which balances downsides and upsides via risk-reward ratios, weighted scoring, and scenario simulations. Finally, it synthesizes inputs from all layers to deliver the final approval decision.

\textbf{Signaling Game Theory:} Our framework models borrower-lender interactions as a signaling game. Borrowers send signals (credit history, income, loan details), while agents act as receivers, updating beliefs based on signals. Hierarchical signaling enables higher-level agents (senders) to guide lower-level ones (receivers), promoting efficient exploration–exploitation trade-offs. The system converges toward Perfect Bayesian Equilibrium, refining beliefs and balancing risk-reward assessments. This mirrors how lenders dynamically infer creditworthiness from borrower signals. Theoretical analysis is presented at \ref{gametheory}.

\section{Experiments}
\textbf{Dataset: } We use credit scoring dataset based on the German Credit Dataset used in financial risk assessment provided by the TheFinAI where it benchmarks multiple datasets and tasks on various LLMs \cite{xie2024FinBen,xie2023pixiu}. Results on cra-lending dataset are reported in the Appendix \ref{cra-lending}.

\textbf{Models: } Our experiments primarily use GPT \cite{openai2024gpt4technicalreport} family models, specifically \textit{gpt-4o} and \textit{o3-mini}. We consider \textit{o3-mini} to be more effective in reasoning tasks, making it a suitable choice for decision-making and overall assessment within our framework. We also show results using Llama3-70B model.

\subsection{Baselines}
We compare our framework against multiple baselines:

\begin{itemize}
    \item \textbf{Zero shot performance}: Evaluate the input query with zero-shot baseline for comparison.
    \item \textbf{Chain of Thought(CoT)}: To assess reasoning ability, we prompt the model with \textit{``Think step by step"} and analyze its response trace within the CoT framework.
    \item \textbf{Single Agent performing Multiple Tasks}: A single agent is assigned the responsibility of performing all subtasks. 
    \item \textbf{Multi Agent System(OURS)}: We experiment with both homogeneous(same model) and heterogeneous setups(different models).
\end{itemize}

To evaluate the robustness of our proposed hierarchical framework, we introduce the following ablations:

\begin{itemize}
\item \textbf{A single-level architecture with multiple agents:} All agents operate at the same level without a hierarchical structure, independently processing different aspects of the credit assessment task.
\item \textbf{A two-level architecture with multiple agents:} Agents are organized into two layers, where the first layer performs the initial pre-processing and assessment, while the second layer performs risk and reward assessment.
\end{itemize}

\begin{table*}
  \centering
      \caption{Performance metrics comparing various credit assessment approaches}
    \begin{tabular}{l|r|r|r|r}
    \toprule
      Evaluation    & Accuracy & Precision & Recall & F1 Score \\
    \toprule
    Zero Shot (\texttt{gpt4o}) & 45.5\%	 & 33.33\%	 &	67.69\%	 &	44.67\% \\
    Zero Shot (\texttt{o3-mini}) & 44\%	 & 47.73\%	 &	59.43\%	 &	52.94\% \\
    Zero Shot (\texttt{Llama3-70B}) & 41.5\%	 & 63.20\%	 &	27.30\%	 &	38.00\% \\
    \hline
    Chain of Thought (\texttt{gpt-4o}) & 36\%	 & 37.12\%	 &	52.13\%	 &	43.36\% \\
    Single Agent performing multitasks(\texttt{gpt-4o})	& 42.5\%	& 28.79 \% & 64.41\% & 39.79 \% \\
    Single Agent performing multitasks(\texttt{o3-mini})	& 45.5\%	& 43.18 \% & 62.64\% & 51.12 \% \\
    \hline
    MultiAgent(OURS) (\texttt{Llama3-70B} \& \texttt{o3-mini}) & 48.50\%	 & \textbf{67.90\%}	 &	41.70\%	 &	51.70\% \\ 
    MultiAgent(OURS) (\texttt{gpt-4o}) & 51\%	 & 65.18\%	 &	55.3\%	 &	59.84\% \\ 
    MultiAgent(OURS) (\texttt{o3-mini}) & 53.5\%	 & 65.12\%	 &	63.64\%	 &	64.37\% \\ 
    \textbf{MultiAgent(OURS) }(\texttt{\textbf{gpt-4o}} \& \texttt{\textbf{o3-mini}}) & \textbf{60\%}	 & 65.48\%	 &	\textbf{83.33\%	} &	\textbf{73.33\%} \\ 
    \toprule
    \end{tabular}%
  \label{tab:performance}%
\end{table*}%

\section{Results and Discussion}

Table \ref{tab:performance} and \ref{cra-lending-performance} highlights that our \textbf{hierarchical MAS significantly outperforms baselines}. Combining \textit{GPT-4o} and \textit{o3-mini} yields 60\% Accuracy (+15.5\% over Zero-Shot GPT-4o), 83.33\% Recall (+15.64\%), and 73.33\% F1 (+20.39\%), while even MAS with \textit{o3-mini} alone surpasses all non-MAS setups (+9.5\% Accuracy, +13.25\% F1). Baseline methods reveal clear limitations: Zero-shot GPT-4o achieves high Recall (67.69\%) but low Precision (33.33\%), showing over-approval bias. \textit{o3-mini} favors Precision (47.73\%) at Recall’s cost (59.43\%). CoT performs worst overall (36\% Accuracy), suggesting reasoning chains propagate errors in credit tasks. \textbf{Single-agent multitasking proves suboptimal} (GPT-4o: 42.5\% Accuracy, 28.79\% Precision). Conflicting priorities hinder decision quality, whereas MAS cross-validation reduces false positives and balances Precision–Recall trade-offs. Ablations (Table \ref{tab:ablations}) show flat architectures yield 9.23\% lower F1 than hierarchical systems. \textit{\textbf{Division of labor enables specialization, error correction, and refinement}}, with later layers validating earlier assessments. Finally, \textbf{heterogeneous MAS combining GPT-4o’s reasoning with o3-mini’s} efficiency achieves the most robust and balanced predictions, reflected in superior Recall and F1 scores.

\section{Biasness Perspective: Towards Fair Lending} \label{bias_section}
We analyze potential gender and ethnicity biases in our multi-agent credit assessment system. For gender \ref{fig:gender}, accuracy dropped from 65.22\% for male applicants to 58.26\% when gender was switched to female while keeping all other features constant. Out of 115 samples, 14 cases showed differing outcomes solely due to gender, indicating bias. Even with gender removed, accuracy further declined to 51.30\%, suggesting indirect bias from correlated features. Loans approved for female applicants also showed lower confidence scores, though strong attributes like stable employment and positive credit history acted as counterbias factors.

For ethnicity \ref{fig:race}, performance varied across groups. African/Black applicants achieved the highest accuracy (57.5\%) but still fell below ground truth (60\%), while Asian applicants had the lowest (52.5\%, -7.5\% below ground truth). All groups underperformed ground truth recall (83.33\%), indicating reduced ability to identify creditworthy applicants. The Asian approval rate (52.5\%) reached only 87.5\% of the baseline, nearing the disparate impact threshold under the 4/5th rule. Additionally, African/Black applicants showed higher recall (75.76\%) but lower precision (65.36\%), implying approval bias despite higher risk. Overall, results demonstrate that both gender and ethnicity significantly influence system outcomes, reinforcing the need for bias mitigation strategies.

\bibliographystyle{plainnat}
\bibliography{references}

\appendix

\section{Dataset Information}
The results are primarily evaluated on two datasets: 

\begin{itemize}
    \item \textit{german-flare} dataset: 200 test samples of the  datatset.
    \item \textit{cra-lending} dataset: 2690 test samples of the dataset.
\end{itemize}

There are 20 features/attributes(13 categorical, 7 numerical) present for each query in the test samples. The credit assessment classifies individuals as ``good" or ``bad" credit risks using historical customer data. 

\section{Results on cra-lending dataset}\label{cra-lending}

\begin{table*}[h]
  \centering
  \caption{Performance metrics comparing various credit assessment approaches on cra-lending dataset}
  \begin{tabular}{l|r|r|r|r}
    \toprule
    Evaluation    & Accuracy & Precision & Recall & F1 Score \\
    \toprule
    Zero Shot (\texttt{gpt-4o}) & 60.5\% & 88.80\% & 58.60\% & 70.60\% \\
    Zero Shot (\texttt{o3-mini}) & 61\% & 89.60\% & 58.60\% & 70.90\% \\
    \hline
    MultiAgent (OURS) (\texttt{gpt-4o} \& \texttt{o3-mini as expert}) & \textbf{66.67\%} & 87.90\% & \textbf{68.10\%} & \textbf{76.70\%} \\
    \toprule
  \end{tabular}%
  \label{cra-lending-performance}
\end{table*}

\section{Ablations study}

\begin{table*}[h]
  \centering
      \caption{Ablations to evaluate the robustness of our proposed hierarchical framework}
    \begin{tabular}{l|r|r|r|r}
    \toprule
      Evaluation & Accuracy & Precision & Recall & F1 Score \\
    \toprule
    Single-level with multiple agents & 46\%	 & 59.38\%	 &	57.58\%	 &	58.46\% \\
    Two-level with multiple agents & 53.77\%	 & 63.70\%	 &	70.45\%	 &	66.91\% \\
    \toprule
    \end{tabular}%
  \label{tab:ablations}%
\end{table*}%

\section{Biasness Figures}

\begin{figure}[h]
    \centering
    \begin{subfigure}{0.49\linewidth}
        \centering
        \includegraphics[width=\linewidth]{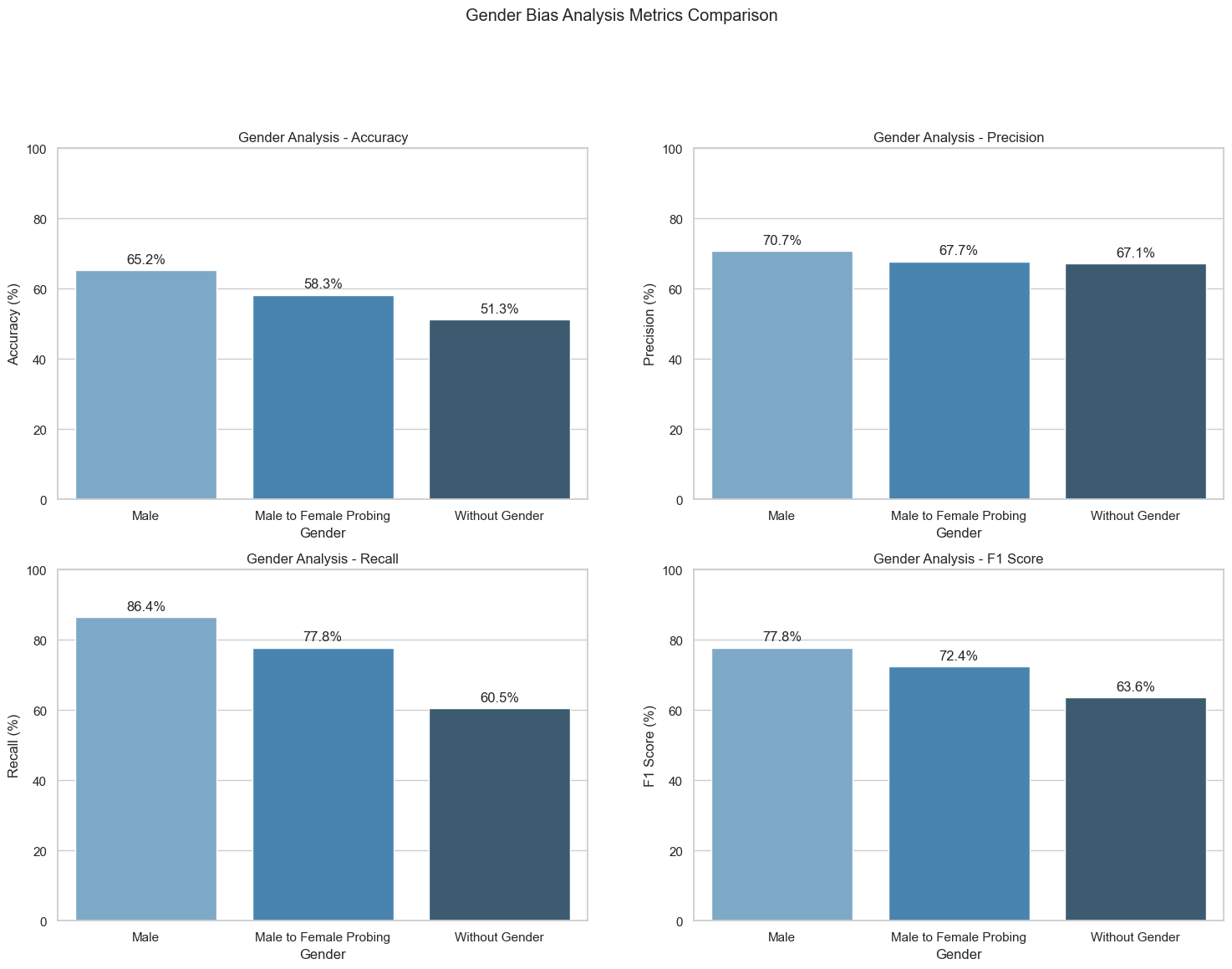}
        \caption{Gender Bias Analysis}
        \label{fig:gender}
    \end{subfigure}
    \hfill
    \begin{subfigure}{0.49\linewidth}
        \centering
        \includegraphics[width=\linewidth]{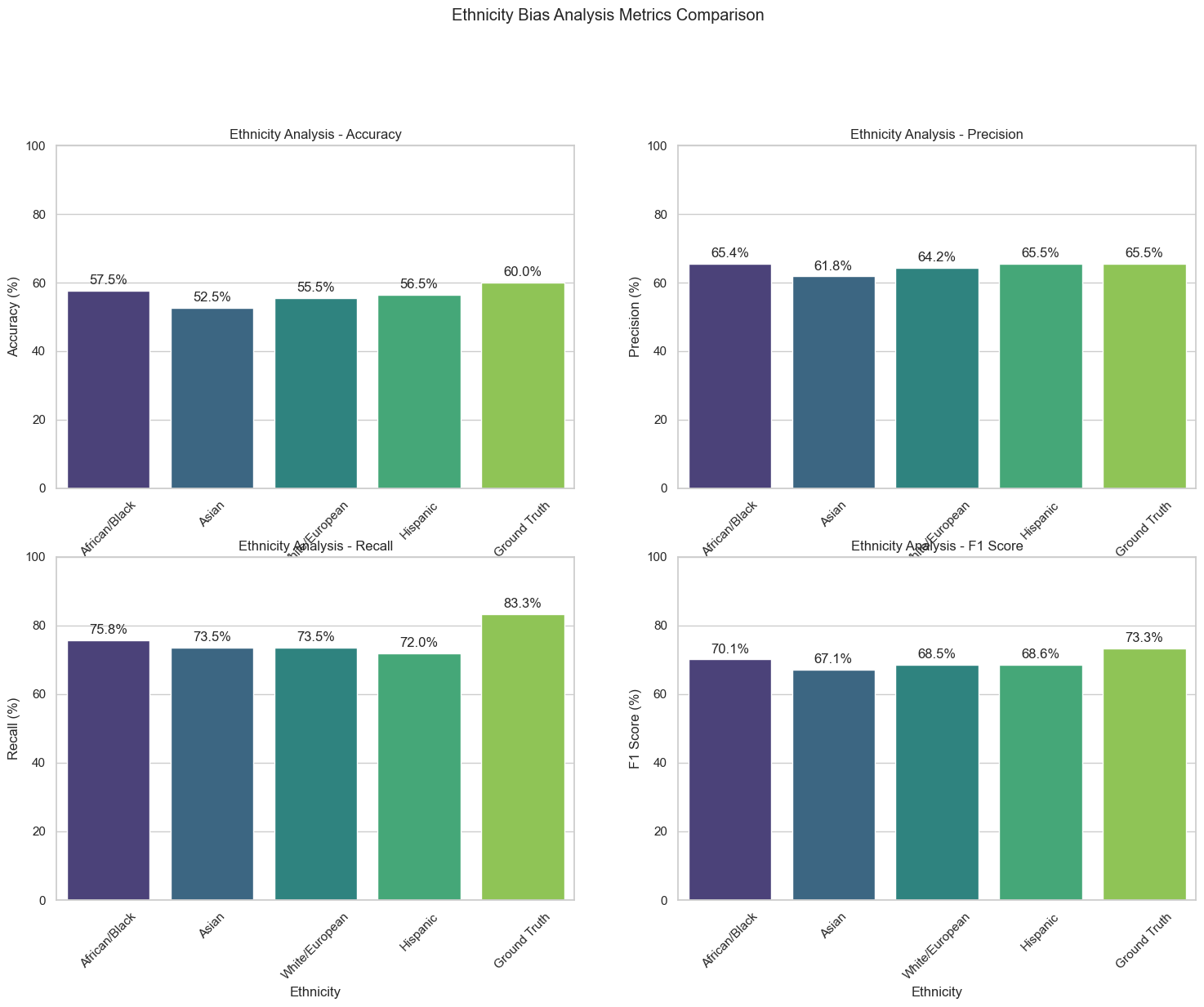}
        \caption{Race Bias Analysis}
        \label{fig:race}
    \end{subfigure}
    \caption{Bias Analysis for Gender and Race}
    \label{bias_fig}
\end{figure}

\section{Signalling Game Theory in Multi Agent Setup}\label{gametheory}

We define our hierarchical MASCA system formally as a signaling game:

$$ (T, M, A, \mu, \sigma, U_S, U_R) $$

where:
\begin{itemize}
    \item $T$: Set of borrower types (creditworthy $t_1$ / risky $t_2$)
    \item $M$: Signals (from initial layer)
    \item $A$: Actions (approve/reject decisions)
    \item $\mu(t|m)$: Receiver's belief about borrower type $t$ given signal $m$
    \item $\sigma(m|t)$: Sender's strategy mapping types to signals
    \item $U_S(t,a)$: Payoff function for sender (borrower)
    \item $U_R(t,a)$: Payoff function for receiver (MAS)
\end{itemize}

\subsection{Mapping to MASCA}
\begin{tabular}{ll}
\toprule
Element & Our Implementation \\
\midrule
Sender & Borrower transmitting financial signals \\
Receiver & MASCA's hierarchical agents analyzing signals \\
Types ($t$) & Borrower creditworthiness: Creditworthy ($t_1$) vs. Risky ($t_2$) \\
Signals ($m$) & Processed outputs from initial layer \\
Actions ($a$) & Approve/Reject decisions by Decision Orchestrator \\
Beliefs ($\mu(t|m)$) & Updated risk probabilities via Bayesian inference \\
\bottomrule
\end{tabular}

\subsection{Sender Strategy}
A borrower of type $t \in \{t_1, t_2\}$ chooses signal $m$ with probability:
$$ \sigma(m|t) = \mathbb{P}(\text{Send } m | \text{Type } t) $$

Example:
\begin{itemize}
    \item Creditworthy borrowers ($t_1$) send ``Strong Credit History'' ($m_1$) with $\sigma(m_1|t_1) = 1$.
    \item Risky borrowers ($t_2$) may mimic $m_1$ with $\sigma(m_1|t_2) = 0.3$.
\end{itemize}

\subsection{Receiver Beliefs}
Posterior probability of borrower type given signal $m$:
$$ \mu(t|m) = \frac{\sigma(m|t) \cdot p(t)}{\sum_{t'} \sigma(m|t') \cdot p(t')} $$
where $p(t)$ is the prior (e.g., 60\% creditworthy, 40\% risky).

\subsection{Receiver Strategy}
Decision Orchestrator chooses action $a \in \{Approve, Reject\}$ to maximize expected utility:
$$ a^*(m) = \arg\max_{a} \mathbb{E}[U_R(t,a)|m] = \sum_{t} \mu(t|m) \cdot U_R(t,a) $$

\subsection{Equilibrium Conditions}
\textbf{Sequential Rationality:}
\begin{itemize}
    \item Senders: $\sigma(m|t)$ maximizes $\mathbb{E}[U_S(t,a)|m]$.
    \item Receivers: $a^*(m)$ optimizes utility given $\mu(t|m)$.
\end{itemize}

\textbf{Belief Consistency:}
Posterior beliefs $\mu(t|m)$ align with sender strategies via Bayes’ rule.

Example: If risky borrowers often mimic $m_1$, then $\mu(t_2|m_1)$ increases, reducing approval rates.

\subsection{Equilibrium Types in MASCA}
\begin{tabular}{lll}
\toprule
Equilibrium Type & MASCA Implementation & Empirical Evidence \\
\midrule
Separating & $t_1$ and $t_2$ send distinct signals & 83.33\% recall [Table 1] \\
Pooling & Both types send identical signals & 9.23\% F1 drop [Table 2] \\
\bottomrule
\end{tabular}

\subsection{Case Study: Employment History Signaling Game}
\textbf{Agents:}
\begin{itemize}
    \item Sender: Contextualizer (Data Ingestion \& Contextualization Layer)
    \item Receiver: Income Stability Analyst (Assessment Layer)
\end{itemize}

\textbf{Payoff Matrix:}

\begin{table}[h!]
\centering
\resizebox{\textwidth}{!}{%
\begin{tabular}{lllll}
\toprule
Sender Type & Signal & Receiver Action & Sender Payoff & Receiver Payoff \\
\midrule
Stable Employment & Low variance & Approve & 5 & 4 \\
Unstable Employment & High variance & Reject & 1 & 5 \\
Unstable Employment & Mimics Low variance & Approve (Error) & 3 & -2 \\
\bottomrule
\end{tabular}%
}
\caption{Signaling game payoff matrix for employment stability and credit approval}
\end{table}

\subsection{Equilibrium Analysis}
\textbf{Separating Equilibrium:}
$$ \sigma(\text{Low}|\text{Stable}) = 1, \quad \sigma(\text{High}|\text{Unstable}) = 1 $$
- Stable applicants truthfully signal low variance $\rightarrow$ Approval.
- Unstable applicants truthfully signal high variance $\rightarrow$ Rejection.

\textbf{Pooling Equilibrium Failure:}
If both types send ``Low variance'':
$$ \mu(\text{Unstable}|m=\text{Low}) = \frac{0.4 \cdot 0.3}{0.6 \cdot 1 + 0.4 \cdot 0.3} = 16.7\% $$

Receiver utility for Approve:
$$ 0.833 \cdot 4 + 0.167 \cdot (-2) = 3.0 < 5 $$
Thus Reject dominates, and pooling equilibrium collapses.

\end{document}